\documentclass{article}

\usepackage[utf8]{inputenc}
\usepackage{scicite}
\usepackage[margin=1in]{geometry}
\usepackage[T1]{fontenc}
\usepackage{url}
\usepackage{booktabs}
\usepackage{amsfonts}
\usepackage{nicematrix}
\usepackage{graphicx}
\usepackage{pgfplots}
\usepackage{hyperref}
\usepackage{float}

\title{ARC Prize 2025: Technical Report}

\author{
    François Chollet,
    Mike Knoop,
    Gregory Kamradt,
    Bryan Landers
}

\begin{document}

\maketitle

\begin{abstract}
 The ARC-AGI benchmark series serves as a critical measure of few-shot generalization on novel tasks, a core aspect of intelligence. The ARC Prize 2025 global competition targeted the newly released ARC-AGI-2 dataset, which features greater task complexity compared to its predecessor. The Kaggle competition attracted 1,455 teams and 15,154 entries, with the top score reaching 24\% on the ARC-AGI-2 private evaluation set. Paper submissions nearly doubled year-over-year to 90 entries, reflecting the growing research interest in fluid intelligence and abstract reasoning. The defining theme of 2025 is the emergence of the  \textit{refinement loop} – a per-task iterative program optimization loop guided by a feedback signal. Refinement loops come in a variety of forms, in particular evolutionary program synthesis approaches and application-layer refinements to commercial AI systems. Such refinement loops are also possible in weight space, as evidenced by zero-pretraining deep learning methods which are now achieving competitive performance with remarkably small networks (7M parameters). In parallel, four frontier AI labs (Anthropic, Google DeepMind, OpenAI, and xAI) reported ARC-AGI performance in public model cards in 2025, establishing ARC-AGI as an industry standard benchmark for AI reasoning. However, our analysis indicates that current frontier AI reasoning performance remains fundamentally constrained to knowledge coverage, giving rise to new forms of benchmark contamination. In this paper, we survey the top-performing methods, examine the role of refinement loops in AGI progress, discuss knowledge-dependent overfitting, and preview ARC-AGI-3, which introduces interactive reasoning challenges that require exploration, planning, memory, goal acquisition, and alignment capabilities.
\end{abstract}

\section{Introduction: ARC-AGI}

In 2019, Fran\c{c}ois Chollet formalized a new definition of artificial general intelligence (AGI), characterizing it as a system capable of efficiently acquiring new skills and solving novel problems for which it was neither explicitly designed nor trained.~\cite{chollet2019intelligence} As a first concrete attempt to measure this new definition of intelligence, Chollet published the Abstraction and Reasoning Corpus (ARC)~\cite{chollet2019github} (later renamed ARC-AGI to avoid naming collisions with other AI benchmarks.) The original benchmark dataset is referred to as ARC-AGI-1. It is a set of independent ``tasks'' (see figure 1), each consisting of a number of ``demonstration pairs'' (two or more, with a median count of three) and one or more ``test inputs''. A test pair consists of an ``input grid'', a rectangular grid of variable size (up to a maximum size of 30 rows by 30 columns) where each cell can have one of ten distinct ``values'', and an output grid which should be fully inferable from the characteristics of the input grid. The goal is to use the demonstration pairs to understand the nature of the task and use this understanding to construct the output grid corresponding to each test input. The test taker is allowed two attempts per test input.

The defining characteristic of the benchmark is that it should not be possible to prepare for any of the tasks in advance. Every task in the dataset follows a different underlying logic, requiring independent rule discovery. All tasks were created by humans to ensure a high degree of novelty and diversity.

ARC-AGI tasks do not require specialized world knowledge (e.g., historical facts) nor language to solve. The only assumed prior knowledge is Core Knowledge~\cite{chollet2019intelligence} – concepts such as objectness, basic topology, elementary integer arithmetic, etc. Human Core Knowledge has been investigated by Spelke et al.~\cite{spelke2007}. These knowledge priors are acquired by children very early (typically before age four) and are universally shared by all humans. The ARC-AGI public training tasks are designed to expose test-takers to all the Core Knowledge priors needed to solve ARC-AGI tasks.

\subsection{Dataset Composition}

Following ARC Prize 2024, we released ARC-AGI-2 in early 2025. ARC-AGI-2 maintains the same task format and core principles as ARC-AGI-1 while incorporating more complex tasks with greater generalization difficulty, and normalizing the distribution of task difficulty.

ARC-AGI-2 consists of a larger set of human-created tasks split into three primary subsets:

\begin{itemize}
    \item Public training tasks (400, imported from ARC-AGI-1) - Intended to demonstrate the task format and allow for learning the Core Knowledge priors.
    \item Semi-Private Evaluation tasks (120) - Used for intermediate leaderboard scoring during competitions. While not publicly released, these tasks have been exposed to commercial APIs and thus carry some risk of leakage.
    \item Private Evaluation tasks (120) - Used for final competition scoring and verification. This set is fully private and theoretically free from leakage.
\end{itemize}

\begin{figure}[h]
    \centering
    \includegraphics[width=1\textwidth]{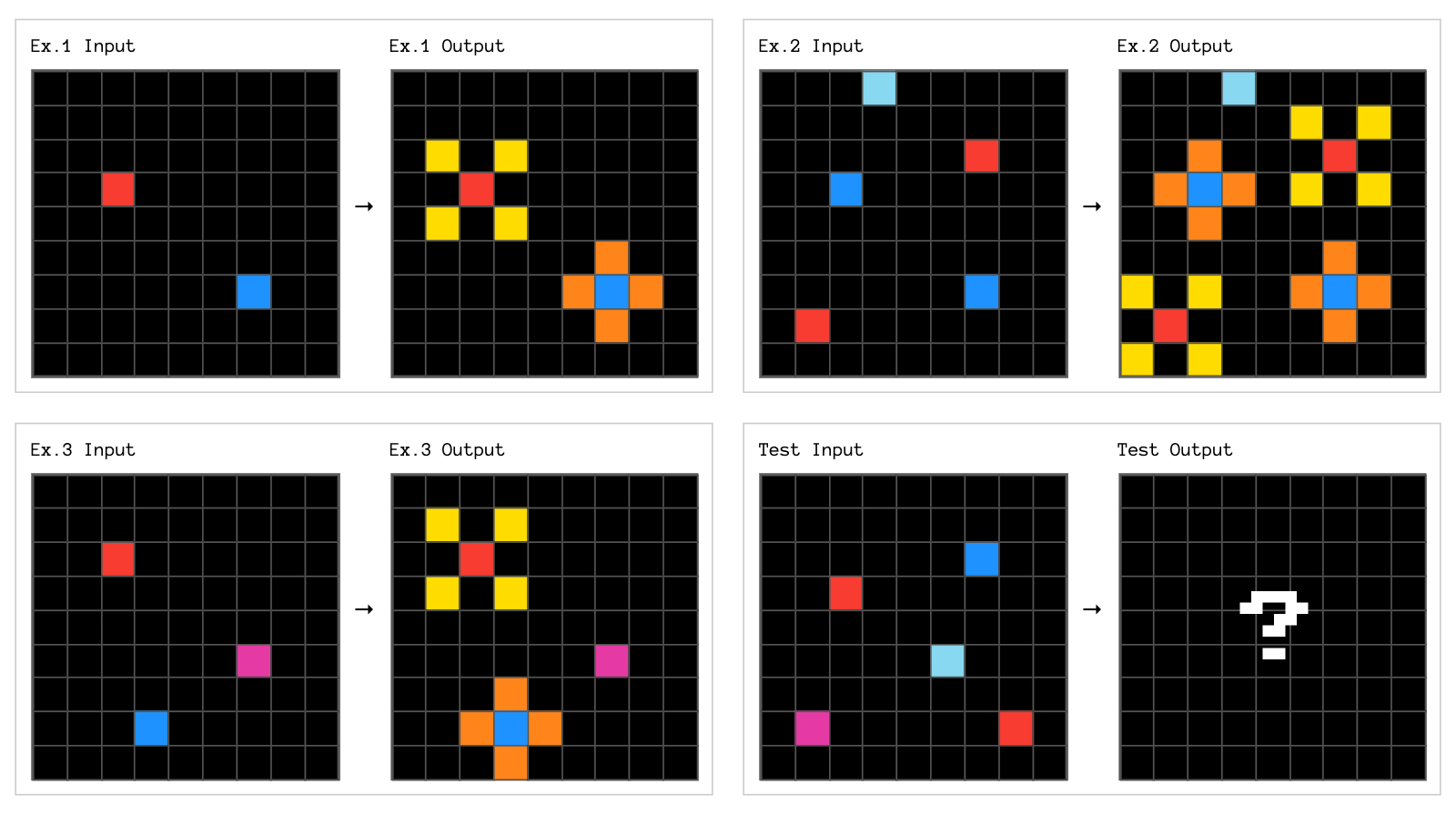}
    \caption{Example ARC-AGI task}
    \label{fig:example-task}
\end{figure}

State-of-the-art scores are only reported on the Semi-Private and Private Evaluation task sets to reduce the risk of overfitting and data contamination.

An important characteristic of ARC-AGI tasks is that they remain hard for AI systems, yet easy for humans. The original ARC-AGI-1 private evaluation tasks were tested by two people who scored 97\% and 98\%, and collectively solved 100\% of tasks.

For ARC-AGI-2, after selection and validation, 100\% of the tasks were solved by at least two (or more) independent, non-expert human testers drawn from the general public, with each task attempted by between 2 and 10 humans. The results of our human study establish that all tasks in ARC-AGI-2 are solvable by humans with no prior training.

\section{ARC Prize 2025 Results}

\subsection{Competition Progress}

ARC Prize 2025 ran from March 26, 2025, to November 3, 2025. In total, 1,455 teams submitted 15,154 entries to the Kaggle competition, at levels similar to ARC Prize 2024. The top competition score reached a new state-of-the-art on the ARC-AGI-2 private dataset of 24\% at a compute cost of \$0.20 per task.

The paper submission track saw significant growth, with 90 papers submitted, up from 47 in 2024. Due to the exceptional quality of submissions, we expanded the paper prizes to include 5 additional runners-up beyond the top three award winners, and recognized 8 additional honorable mentions.

All ARC Prize 2025 winning solutions and papers are open-source and available at \href{https://arcprize.org}{arcprize.org}.

\subsection{Top Scores}
\begin{table}[h]
  \centering
  \small
  \begin{tabular}{lllr}
    \toprule
    \textbf{Place} & \textbf{Prize} & \textbf{Team} & \textbf{ARC-AGI-2 Private Score} \\
    \midrule
    1st & \$25k & NVARC \cite{nvarc2025} & 24.03\% \\
    2nd & \$10k & the ARChitects \cite{architects2025} & 16.53\% \\
    3rd & \$5k & MindsAI \cite{mindsai2025} & 12.64\% \\
    4th & \$5k & Lonnie \cite{lonnie2025} & 6.67\% \\
    5th & \$5k & G. Barbadillo \cite{barbadillo2025} & 6.53\% \\
    \bottomrule
  \end{tabular}
  \caption{ARC Prize 2025 Top Score winners.}
\end{table}

The top three Kaggle submissions demonstrate continued progress in test-time training and ensemble techniques.

\textbf{1st Place - NVARC (24.03\%):} This entry builds upon the 2024 ARChitects winning entry (which leverages test-time training) and makes heavy use of synthetic data generation to improve model performance.

\textbf{2nd Place - the ARChitects (16.53\%):} A 2D-aware, masked-diffusion language model with recursive self-refinement and perspective-based scoring. This solution improved substantially over the team's 2024 autoregressive system through novel architectural modifications tailored for spatial reasoning.

\textbf{3rd Place - MindsAI (12.64\%):} A heavily-engineered test-time-training pipeline that combines test-time fine-tuning, augmentation ensembles, tokenizer dropout, and novel pretraining techniques to produce a competitive score on ARC-AGI-2.

Video interviews with the 1st, 2nd, and 3rd place Top Score winners are available on the ARC Prize website~\cite{arcprize2025_videos}.

\subsection{Paper Awards}
\begin{table}[H]
  \centering
  \small
  \begin{tabular}{llp{4.1cm}p{7.0cm}}
    \toprule
    \textbf{Place} & \textbf{Prize} & \textbf{Authors} & \textbf{Title} \\
    \midrule
    1st & \$50k & A. Jolicoeur-Martineau & \textit{Less is More: Recursive Reasoning with Tiny Networks} \cite{jolicoeur2025less} \\
    2nd & \$20k & J. Pourcel, C. Colas, P. Oudeyer & \textit{Self-Improving Language Models for Evolutionary Program Synthesis: A Case Study on ARC-AGI} \cite{pourcel2025selfimproving} \\
    3rd & \$5k & I. Liao, A. Gu & \textit{ARC-AGI Without Pretraining} \cite{liao2025arcagi} \\
    Runner Up & \$2.5k & I. Joffe, C. Eliasmith & \textit{Vector Symbolic Algebras for the Abstraction and Reasoning Corpus} \cite{joffe2025vsa} \\
    Runner Up & \$2.5k & J. Berman & \textit{From Parrots to Von Neumanns: How Evolutionary Test-Time Compute Achieved State-of-the-Art on ARC-AGI} \cite{berman2025parrots} \\
    Runner Up & \$2.5k & E. Pang & \textit{Efficient Evolutionary Program Synthesis} \cite{pang2025efficient} \\
    Runner Up & \$2.5k & E. Guichard, F. Reimers, M. Kvalsund, M. Lepperod, S. Nichele & \textit{ARC-NCA: Towards Developmental Solutions to the Abstraction and Reasoning Corpus} \cite{guichard2025arcnca} \\
    \bottomrule
  \end{tabular}
  \caption{ARC Prize 2025 Paper Award winners.}
\end{table}

The top three Paper Awards recognized novel approaches that advance the theoretical and practical understanding of artificial fluid intelligence.

\textbf{1st Place - Jolicoeur-Martineau:} The Tiny Recursive Model (TRM) is a 7M-parameter, single-network recursive model with separate answer and latent states. Using deep supervised refinement, TRM demonstrates that extremely small networks can achieve competitive ARC-AGI performance when trained with appropriate recursive reasoning mechanisms.

\textbf{2nd Place - Pourcel, Colas, and Oudeyer:} SOAR (Self-improving Operators for Automated program Refinements) is a self-improving evolutionary program synthesis framework that fine-tunes an LLM on its own search traces. This approach improves open-source ARC-AGI-1 solution performance by up to 52\% without requiring human-engineered domain-specific languages (DSLs) or solution datasets, demonstrating the potential for autonomous improvement in program synthesis systems.

\textbf{3rd Place - Liao and Gu:} CompressARC is an MDL-based (Minimum Description Length), single puzzle-trained neural code golf system that achieves 20-34\% on ARC-AGI-1 and 4\% on ARC-AGI-2 without any pretraining or external data. This work demonstrates that pure test-time optimization based on description length minimization can solve abstract reasoning tasks without leveraging large-scale pretraining.

Video interviews with the Paper Award winners are available on the ARC Prize website  channel~\cite{arcprize2025_videos}.

\subsection{Honorable Mentions}

Eight additional papers received honorable mention recognition for their contributions to ARC-AGI research:

\begin{itemize}
    \item K. Hu et al., \textit{``ARC-AGI is a Vision Problem!''} \cite{hu2025vision}
    \item D. Franzen, J. Disselhoff, D. Hartmann, \textit{``Product of Experts with LLMs: Boosting Performance on ARC Is a Matter of Perspective''} \cite{franzen2025product}
    \item G. Barbadillo, \textit{``Exploring the combination of search and learn for the ARC25 challenge''} \cite{barbadillo2025}
    \item A. Das, O. Ghugarkar, V. Bhat, J. McAuley, \textit{``Beyond Brute Force: A Neuro-Symbolic Architecture for Compositional Reasoning in ARC-AGI-2''} \cite{das2025beyond}
    \item R. McGovern, \textit{``Test-time Adaptation of Tiny Recursive Models''} \cite{mcgovern2025testtime}
    \item P. Acuaviva et al., \textit{``Rethinking Visual Intelligence: Insights from Video Pretraining''} \cite{acuaviva2025rethinking}
    \item J. Cole, M. Osman, \textit{``Don't throw the baby out with the bathwater: How and why deep learning for ARC''} \cite{mindsai2025}
    \item I. Sorokin, J. Puget, \textit{``NVARC solution to ARC-AGI-2 2025''} \cite{nvarc2025}
\end{itemize}

\section{Program Refinement Loops}

The central theme driving AGI progress in 2025 is the emergence of the refinement loop. At its core, a refinement loop iteratively transforms one program or model version into a slightly better one, based on a feedback signal.

\subsection{Types of refinement loops}

Representatives of refinement loops include:

\begin{itemize}
    \item \textbf{Deep learning with test-time training methods}, where the program being refined is the weights of a pretrained model.
    \item \textbf{Zero-pretraining deep learning methods} such as TRM.
    \item \textbf{Evolutionary Program Synthesis} in either symbolic program space or natural language program space.
    \item \textbf{Test-time Chain-of-Thought optimization} with feedback from a verifier model.
\end{itemize}

Among these, two represent especially interesting developments for 2025: Evolutionary Program Synthesis and Zero-pretraining deep learning methods.

\subsubsection{Evolutionary Program Synthesis}

Examples of this technique include J. Berman~\cite{berman2025parrots} and E. Pang~\cite{pang2025efficient}. Berman's approach employs an evolutionary search harness that evolves ARC solution programs in natural language. Pang's approach follows a similar strategy but operates in Python and dynamically creates a program abstraction library to guide synthesis.

Both approaches implement a two-phase refinement process. First, an exploration phase generates many candidate solutions. Second, a verification phase analyzes these programs to produce a feedback signal. This cycle repeats per task until the resulting program is fully refined and provides accurate answers for all training input/output pairs.

\subsubsection{Zero-Pretraining Deep Learning Methods}

Refinement loops are also becoming the basis for a new type of training paradigm for deep learning models.

Classically, deep learning models are trained on input/output pairs using gradient descent to create a static neural network. This training algorithm gradually refines a high-dimensional curve in the network's latent space. At inference time, when presented with a new input, the network performs forward passes to approximate the output based on this curve. In 2023 and 2024, this paradigm was expanded to add test-time training, where the network is further fine-tuned at inference time on examples of a novel task (a process which is itself a kind of program refinement loop). In this paradigm, the neural weights are trained to represent a task-specific solver program. Test-time training is responsible for the top scores in both ARC Prize 2024 (the ARChitects) and 2025 (NVARC).

Input/output pairs are still used as ground truth and refinement loops play a key role in training, mirroring program synthesis approaches but operating in weight space rather than symbolic space.

More recent work demonstrates early success with a version of this that does away with pretraining entirely, initializing a network from scratch for a specific task and fitting the weights to encode a task-solving program using only examples of that task. Examples include Liao et al.~\cite{liao2025arcagi}, Hierarchical Reasoning Models (HRM)~\cite{wang2025hierarchical} and Jolicoeur-Martineau~\cite{jolicoeur2025less}.

This approach exhibits two unusual properties:

\begin{enumerate}
    \item The resulting networks are extremely small relative to their ARC-AGI performance.
    \item All material task-specific training occurs at test time.
\end{enumerate}

\subsection{Open Source Examples}

\subsubsection{Tiny Recursive Model (TRM)}

\begin{figure}[h]
    \centering
    \includegraphics[width=0.4\textwidth]{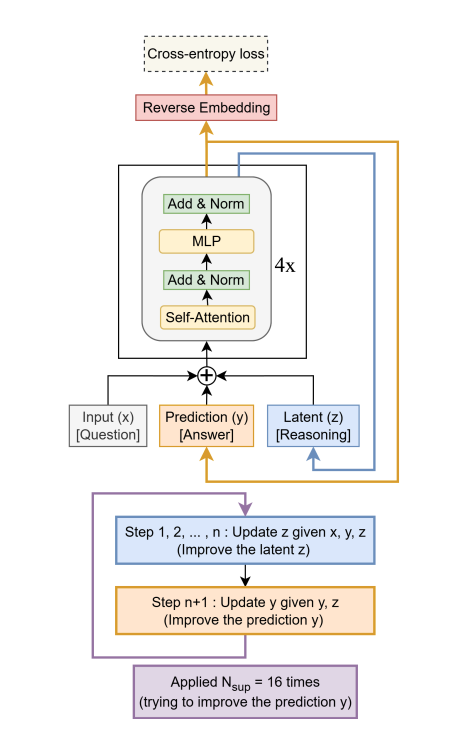}
    \caption{TRM architecture.}
    \label{fig:trm-architecture}
\end{figure}

The Tiny Recursive Model (TRM) (Paper Award 1st Place, Jolicoeur-Martineau~\cite{jolicoeur2025less}), which builds upon the earlier Hierarchical Reasoning Model (HRM)~\cite{wang2025hierarchical}, achieves 45\% test accuracy on ARC-AGI-1 and 8\% on ARC-AGI-2 with only a 7M parameter network. From the paper:

\begin{quote}
Tiny Recursive Model (TRM) recursively improves its predicted answer $y$ with a tiny network. It starts with the embedded input question $x$ and initial embedded answer $y$, and latent $z$. For up to $N_{sup} = 16$ improvement steps, it tries to improve its answer $y$. It does so by i) recursively updating $n$ times its latent $z$ given the question $x$, current answer $y$, and current latent $z$ (recursive reasoning), and then ii) updating its answer $y$ given the current answer $y$ and current latent $z$. This recursive process allows the model to progressively improve its answer (potentially addressing any errors from its previous answer) in an extremely parameter-efficient manner while minimizing overfitting.
\end{quote}

\subsubsection{CompressARC}

\begin{figure}[h]
    \centering
    \includegraphics[width=0.9\textwidth]{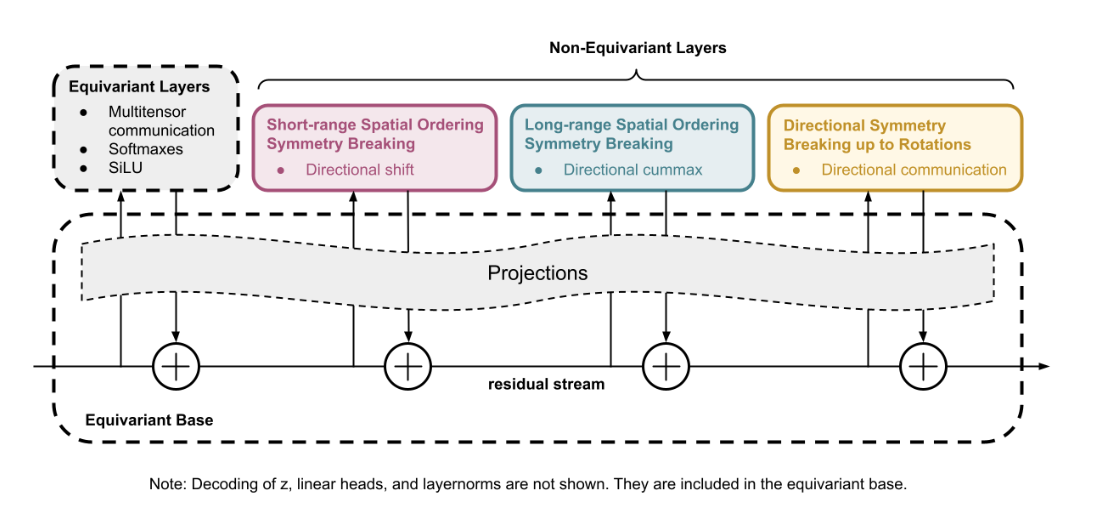}
    \caption{CompressARC architecture and approach.}
    \label{fig:compressarc}
\end{figure}

CompressARC (Paper Award 3rd Place, Liao and Gu~\cite{liao2025arcagi}) uses only 76K parameters, yet achieves 20\% on the ARC-AGI-1 evaluation set, processing each puzzle in approximately 20 minutes on a single NVIDIA RTX 4070 GPU.

This solution features three distinctive characteristics:

\begin{itemize}
    \item \textbf{No pretraining:} Models are randomly initialized and trained only at test time.
    \item \textbf{No dataset:} One model trains on a single target task and produces one answer.
    \item \textbf{No branching search:} The approach relies solely on gradient descent.
\end{itemize}

The method operates by minimizing the description length of each task at test time, following the Minimum Description Length (MDL) principle. Liao derives how a standard Variational Autoencoder (VAE) loss with decoder regularization can substitute for combinatorial search to refine very small neural network programs. The generalization achieved by such a compact network is remarkable.

\subsection{Commercial Examples}

Evidence of iterative refinement appears in commercial AI reasoning systems. A Chain-of-Thought can be interpreted as a natural language program that transforms one latent state into another.

Consider ARC-AGI-1 task \textbf{\#4cd1b7b2}. Gemini 3 Pro used 96 reasoning tokens to solve this task, whereas Gemini 3 Deep Think employed 138,000 tokens. Higher reasoning modes for these systems exhibit a strong correlation with increased reasoning token counts (longer programs), even when not strictly necessary for task completion.

These extended natural language programs enable more refinement through additional exploration and verification cycles. Analysis of reasoning outputs from commercial systems reveals self-corrective behavior:

\begin{quote}
\textit{... which fails the complete set requirement. This suggests the current solution might not fully satisfy the puzzle's constraints. I need to re-examine the box configuration and explore alternative arrangements ...} (Claude Opus 4.5)
\end{quote}

\begin{quote}
\textit{... which suggests further investigation is needed to complete the analysis. I'll verify the center point at row 9, column 15 ...} (Claude Opus 4.5)
\end{quote}

\begin{quote}
\textit{... maybe each input row is being duplicated three times in the output, but how does that fit with the rest? Wait, the third output row is ...} (QwQ 32B)
\end{quote}

An important finding with frontier commercial models released in late 2025 (Gemini 3, Claude Opus 4.5, etc.) is that refinement loops can be implemented at the application layer to meaningfully improve task reliability, rather than relying solely on provider reasoning systems. This approach still requires that the foundational model possess adequate knowledge coverage of the task domain.

\subsection{Model Refinement Harnesses}

In the last quarter of 2025, we introduced a new leaderboard category termed ``Model Refinements'' to track application-layer improvements to commercial AI systems, also known as harnesses. We verified a Gemini 3 Pro refinement harness implementation open-sourced by Poetiq, which improves performance on ARC-AGI-2 from a baseline of 31\% accuracy at \$0.81 per task to 54\% accuracy at \$31 per task. The same refinement approach achieved comparable gains on Claude Opus 4.5, with accuracy rivaling Gemini 3 Pro but at about twice the cost per task (approximately \$60 per task), as reported by Poetiq.

Currently, the refinement harnesses we observe are domain-specific. However, techniques such as GEPA~\cite{gepa2025} and DSPy~\cite{khattab2024dspy} enable the development of general-purpose reliability improvements at the application layer, provided a verifier or environment capable of producing a feedback signal is available.

We anticipate that general refinement harness improvements will eventually be integrated behind the API of commercial AI systems. Simultaneously, we expect that bleeding-edge, task-specific accuracy will continue to be driven by knowledge specialization and application-layer verifiers.

\section{AGI Progress \& The Future of ARC-AGI}

As of 2025, with the advent of AI reasoning systems, task domains with the following two characteristics are reliably automatable, with no new science needed:

\begin{enumerate}
    \item Sufficient task knowledge coverage exists in the pretraining corpus.
    \item The task provides a verifiable feedback signal.
\end{enumerate}

Current AI reasoning performance is fundamentally related to model knowledge. This relationship warrants careful consideration, as human reasoning capability is not similarly bound to knowledge. This coupling has various implications and leads to imprecise characterizations such as ``jagged intelligence.''~\cite{karpathy2024jagged}

Supporting evidence for this knowledge-dependent reasoning emerged across multiple domains in 2025, including performance on ARC-AGI-2 (abstract reasoning), 2025 IMO Gold Medal achievement (mathematics), and 2025 ICPC 100\% performance (competitive programming) – all driven by AI reasoning systems. These task domains are substantially broader than those addressable by pure language models without reasoning capabilities. However, they remain relatively narrow in a global context.

The invention and scaling of chain-of-thought synthesis represents a profound upgrade in AI capability comparable to the invention and scaling of transformers. However, we are still in the early stages of deployment. Few users have directly experienced these tools. According to Sam Altman (OpenAI), approximately 7\% of ChatGPT free users have engaged with ``thinking’’ mode~\cite{altman2025thinkingmode}. We expect diffusion of current technology to require 5-10 additional years, even within business contexts alone.

Collecting domain knowledge and building verifiers is not cost-free. This represents relatively expensive and specialized work. Presently, AI automation is a function of the societal willingness to invest in the necessary talent, compute, and data resources. We anticipate steady progress over the next 12-24 months as society conducts a global search for problems that are both (1) most important and (2) fall within acceptable cost thresholds. This includes early results in which AI systems produce novel scientific knowledge in fields with adequate knowledge coverage. Recent work by Hsu reports an AI refinement loop using a generator–verifier architecture to produce novel results in quantum physics~\cite{hsu2025quantumrefinement}.

However, many potentially automatable problems fall beyond current societal cost cutoffs. As engineering advances, costs will decrease, expanding the set of domains that can be automated. More broadly, machines capable of highly efficient adaptation to produce paradigm-shifting innovation remain firmly within the realm of science fiction.

For the ARC-AGI-1 and ARC-AGI-2 format, we assess that the Grand Prize accuracy gap is now primarily bottlenecked by engineering, while the efficiency gap remains bottlenecked by fundamental science and new ideas. ARC Prize exists to inspire and reward open AGI progress, and as previously committed, we will continue operating the ARC-AGI-2 Grand Prize competition in 2026 to track progress toward a fully open and reproducible solution.

Despite their capabilities, AI reasoning systems still exhibit numerous flaws and inefficiencies necessary to overcome in order to reach AGI. We still need new ideas, such as methods to separate knowledge and reasoning, among other challenges. New benchmarks will be needed to highlight the moment in which those ideas arrive.

\subsection{Knowledge Overfitting}

In machine learning, overfitting occurs when a model learns excessive detail from training data. The model memorizes exact training examples rather than learning general patterns, leading to poor performance on unseen test data. A common AI benchmarking critique is that model providers are incentivized to ``benchmark maximize'' or ``train to the test'' to report high scores for marketing purposes that do not generalize to real-world applications. ARC-AGI-1 and ARC-AGI-2 were designed to resist this style of overfitting by employing a private dataset for official scoring and verification.

AI reasoning systems have altered the landscape in a manner that reflects genuine progress. They have demonstrated non-zero fluid intelligence and can adapt to tasks somewhat removed from their precise knowledge base when the foundational model is grounded in the broader domain. This means that even well-designed benchmarks resistant to direct memorization can now be ``overfit'' if the public training and private test sets are too similar (e.g., independent and identically distributed) and the model has been trained on substantial public domain data.

We assert that this phenomenon is now occurring with ARC-AGI-1 and ARC-AGI-2 – accidentally or intentionally, although we cannot determine which.

Evidence from our Gemini 3 verification demonstrates this pattern:

\begin{quote}
\textit{... Target is Green (3). Pattern is Magenta (6) Solid. Result: Magenta Square on Green ...} (Gemini 3 Deep Think)
\end{quote}

Our LLM verification harness does not mention ARC-AGI tasks or color formats, yet the model employs correct ARC color mappings in its reasoning. This strongly suggests that ARC data are well-represented in the underlying model – sufficiently so to make correct ARC inferences based solely on the structure and format of 2D JSON arrays of integers.

\subsection{Characterizing AGI through continual benchmark adaptation}

Although we assess that this new form of ``overfitting'' assists models in solving ARC, we cannot precisely quantify the magnitude of this effect. Regardless, the ARC-AGI-1 and ARC-AGI-2 formats have provided a valuable scientific indicator of AI reasoning progress. However, benchmark design must adapt.

In fact, ARC Prize has revealed a broader lesson over the past two years: the most valuable and effective benchmarks are created by teams fundamentally committed to driving progress. Such progress requires a sustained dedication to understanding the underlying technology through serious study, a willingness to identify flaws and incentivize corrective action, and adaptation as the technology improves. It also requires a year-over-year commitment. Building effective benchmarks demands sustained effort.

The critical concept is \textit{adaptation}. Adaptation represents the core mode of intelligence. This process extends beyond creating effective benchmarks – it constitutes the ultimate measure of general intelligence itself.

From François Chollet in December 2024~\cite{chollet2024agiquote}:

\begin{quote}
\textit{You'll know AGI is here when the exercise of creating tasks that are easy for regular humans but hard for AI becomes simply impossible.}
\end{quote}

This captures the ARC-AGI benchmark design methodology: operate a real world refinement loop by iteratively improving benchmarks in response to AI progress to drive the gap between ``easy for humans, hard for AI'' toward zero.

By this definition, we have not yet achieved AGI. We are actively developing ARC-AGI-3 for release in early 2026 and are optimistic about the new format. We anticipate that it will stimulate the development of entirely novel ideas.

\subsection{ARC-AGI-3}

Over the past six months, we have focused on developing ARC-AGI-3. Like all versions of ARC, it is designed to be ``easy for humans, hard for AI,'' while serving as the most valuable and scientifically useful benchmark pointing toward AGI and identifying what remains necessary to achieve it. We are building hundreds of never-before-seen interactive environments designed to test agentic reasoning.

We plan to release ARC-AGI-3 in early 2026 alongside ARC Prize 2026. This new benchmark version marks the first major format change since ARC-AGI was introduced in 2019. While the first two versions challenged static reasoning, the third version is designed to challenge interactive reasoning and requires new AI capabilities to succeed: exploration, planning, memory, goal acquisition, and alignment.

Efficiency is a fundamental aspect in the measurement of intelligence. We are particularly optimistic that the ARC-AGI-3 scoring metric will enable formal comparison of human and AI action efficiency (i.e., learning efficiency) for the first time.

\section{Conclusions}

ARC Prize 2025 demonstrated continued open-source progress towards AGI, with the top Kaggle score reaching 24\% on ARC-AGI-2 and significant growth in paper submissions (90, up from 47 in 2024). The emergence of the refinement loop as a central theme represents a significant shift in approaches to abstract reasoning, evidenced by both zero-pretraining deep learning methods and commercial AI reasoning systems.

The year revealed that AI reasoning performance remains fundamentally constrained by knowledge coverage, a characteristic distinct from human reasoning which is capable of extreme generalization. While this enables reliable automation of tasks with sufficient knowledge coverage and verifiable feedback signals, it also introduces new forms of overfitting that require adaptive responses in benchmark design.

ARC-AGI has served as a valuable scientific indicator for AI reasoning progress, with four major AI labs (Anthropic, Google DeepMind, OpenAI, and xAI) reporting ARC-AGI performance on model cards. However, the benchmark must continue to evolve. The development of ARC-AGI-3, scheduled for release in early 2026, represents this commitment to adaptation -- the core principle of intelligence itself.

We remain committed to operating ARC Prize annually until the benchmark is defeated with a public reference solution. All ARC Prize 2025 winning solutions and papers are open-source, reproducible, and available at \href{https://arcprize.org}{arcprize.org}. We aim for ARC-AGI to continue serving as a focal point for research on generalization and reasoning, and to support sustained open progress towards AGI.

\section{Appendix}

\subsection{Acknowledgments}

ARC Prize 2025 builds on the foundation established by previous ARC-AGI competitions, including the 2020 Kaggle competition and the 2022-2023 ARCathons. We are grateful for the continued evolution of the benchmark and the growing community it has fostered.

ARC Prize would not be possible without the full support of the ARC Prize team, our competition partners at Kaggle, and our sponsors. Kaggle plays a critical role in the artificial intelligence and machine learning ecosystem, and their continued partnership has been instrumental in the success of ARC Prize 2025.

We extend our gratitude to all frontier AI labs who worked with us in 2025 to verify their new AI systems on ARC-AGI, including OpenAI, xAI, Anthropic, and Google DeepMind. This collaboration has been invaluable in establishing ARC-AGI as a meaningful benchmark for reasoning progress.

We recognize the dedication of our community members who have built tools, answered questions, and served as resources for researchers and participants throughout the year.

Finally, we extend our deepest gratitude to all participants in ARC Prize 2025, especially those who shared their work openly with the community. Your dedication advances the broader field of AI, bringing us closer to realizing the transformative potential of AGI for humanity. We are inspired by everyone with new ideas who works on ARC-AGI and remain committed to stewarding this attention as a north star toward AGI.

\bibliographystyle{plain}
\bibliography{references}

\end{document}